\documentclass[runningheads]{llncs}
\usepackage[T1]{fontenc}
\usepackage{graphicx}
\usepackage{booktabs}
\usepackage[misc]{ifsym}
\usepackage[hidelinks]{hyperref}
\newcommand{\corr}{(\Letter)}
\usepackage{booktabs} 
\usepackage{multirow} 
\usepackage{tikz}  
\usetikzlibrary{positioning, arrows.meta, fit}  
\usepackage{mwe}
\usepackage{soul}
\usepackage{xcolor}

\usepackage{xcolor} 
\usepackage{inconsolata} 
\usetikzlibrary{fit,positioning,calc,arrows.meta,backgrounds} 

\definecolor{panelbg}{HTML}{F5F5F5} 
\definecolor{jsonkey}{HTML}{1A1A1A} 
\definecolor{jsonstr}{HTML}{0B6E4F} 
\definecolor{jsonnum}{HTML}{B23A00} 
\definecolor{blueann}{HTML}{1F6FEB} 
\definecolor{redann}{HTML}{C81E1E} 

\newcommand{\jk}[1]{{\color{jsonkey}#1}} 
\newcommand{\js}[1]{{\color{jsonstr}#1}} 
\newcommand{\jn}[1]{{\color{jsonnum}#1}} 

\usepackage{mdframed} 
\usepackage{colortbl} 
\usepackage{array}

\begin{document}
\title{Schema-Constrained Document-Level Event Argument Extraction with Lightweight LLM Fine-Tuning}


\titlerunning{Document-Level Event Argument Extraction with LLM Fine-Tuning}


\authorrunning{R. Pietrantuono et al.}




\author{Roberto Pietrantuono\inst{1}\corr \and
Antonio Guerriero\inst{2} \and
Pouya Sattari\inst{1}}

\institute{University of Naples Federico II, Naples, Italy\\
\email{roberto.pietrantuono@unina.it, p.sattari@studenti.unina.it}
\and
University of Salerno, Fisciano, Italy\\
\email{antguerriero@unisa.it}}

\tocauthor{Roberto Pietrantuono, Antonio Guerriero, Pouya Sattari}
\toctitle{Schema-Constrained Document-Level Event Argument Extraction with Lightweight LLM Fine-Tuning}

\maketitle              

\begin{abstract}

Event Argument Extraction (EAE) converts documents into structured event records by identifying argument spans and assigning them schema-defined roles. Document-level EAE is challenging due to long-range dependencies between triggers and arguments, cross-sentence context, and strict role constraints, which often lead to boundary errors, uncertainty in roles, and inconsistencies with restricted schemas.

In this paper, we study whether mid-sized open LLMs can perform schema-constrained EAE reliably at the document level on \texttt{MAVEN-ARG}. Our approach combines (i) role-set injection in prompts for schema compliance, (ii) parameter-efficient supervised fine-tuning (LoRA) using the same JSON-only interface used at inference, and (iii) deterministic decoding with post-processing that validates JSON, filters invalid roles, de-duplicates arguments, and aligns spans to the document window. Under the official \texttt{MAVEN-ARG} evaluator, fine-tuned \textit{mid-sized open models} outperform previously reported GPT baselines across mention, entity-coreference, and event-coreference evaluations; our best model (Phi-4, 14B) reaches 42.39\% F1 at the event-coreference level. Code to reproduce experiments is publicly available at \url{https://github.com/dessertlab/EAE/}.

\end{abstract}

\section{Introduction}
Event Argument Extraction (EAE) extends event extraction beyond detecting triggers and event types to recovering who did what to whom, when, where, and how by identifying \textit{argument spans} and assigning them to schema-defined roles. EAE is a key step for converting unstructured documents into structured event records for downstream applications such as timeline construction, analytics, and information retrieval. 
Figure~\ref{fig:eae-graph} provides a high-level view of document-level event understanding. The sentence contains two \textbf{Event Type}s (\emph{Military Operation} and \emph{Loss}), each anchored by a \textbf{Trigger} in the text (e.g., \emph{invasion} for \emph{Military Operation} and \emph{lost} for \emph{Loss}). For each event, the goal of EAE is to extract an \textbf{Argument} span from the document and assign it an \textbf{Argument Role} from the event schema (e.g., \emph{Agent}, \emph{Patient}, \emph{Location}, \emph{Content}) as illustrated by the dashed links. 
The figure also illustrates cross-event relations (e.g., \textsc{BEFORE} and \textsc{CAUSE}) between event types, motivating document-level context. In this work, we focus on schema-constrained EAE: given a document, an event type, and a marked trigger, we predict role-to-argument spans as valid JSON that follows the role schema and can be evaluated by the official scorer. 

Prior work on EAE has progressed from rule-based systems \cite{bui-etal-2013-fast} to traditional machine learning approaches \cite{Kalimoldayev2023MAPPINGTS,valenzuela-escarcega-etal-2015-domain} and deep learning models – such as CNNs, RNNs, transformer architectures – to better capture context and semantic relations \cite{Li2021ACS}. 
More recently, large language models (LLMs) have been explored for event extraction via prompting and few-shot learning, offering flexibility across domains and languages \cite{Rollo2024LeveragingLF}. 

However, document-level EAE remains challenging because correct outputs must satisfy two constraints simultaneously. First, predictions must follow a \textbf{strict role schema}: for a given event type, only a fixed set of roles is valid, and models must avoid producing extra keys or mismatched roles. Second, extracted arguments must be \textbf{span-faithful}: argument strings must correspond to substrings of the document, even when supporting evidence is distributed across multiple sentences or far from the trigger. These constraints amplify common errors such as boundary mismatches, role confusion in long-tailed inventories, and invalid or inconsistent structured outputs, especially when LLM generations drift into free-form text.

In this work, we investigate whether \emph{mid-sized open LLMs} can perform \emph{schema-constrained document-level EAE} reliably on MAVEN-ARG, a benchmark with a large ontology and coreference-aware evaluation.
A central question we address is whether lightweight LLMs, when adapted with a strict schema-conditioned interface, can achieve competitive document-level EAE performance under the official MAVEN-ARG evaluation protocol.
Our goal is not to propose a new model architecture, but to provide a \emph{reproducible end-to-end pipeline} that produces structured outputs in the exact format required by the official evaluator. The contribution is therefore deliberately empirical and systems oriented, since it offers a controlled single GPU comparison of mid-sized open LLMs under one schema-conditioned interface and the official scorer, rather than a new modeling paradigm. The pipeline uses a JSON-only instruction format, injects the valid role set for each event type into the prompt to constrain outputs, applies parameter-efficient supervised fine-tuning to adapt open models to the task, and performs deterministic validation to enforce schema consistency and span faithfulness before evaluation. We summarize our contributions as follows:

\begin{figure}[t]
\includegraphics[width=0.85\textwidth]{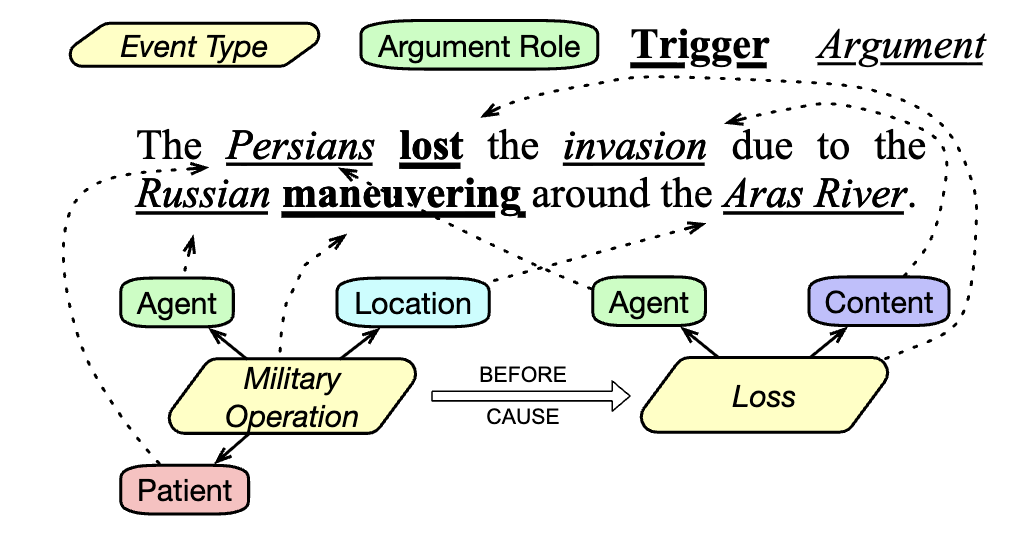}
\caption{An example of the EAE problem \cite{wang-etal-2024-maven}} \label{fig:eae-graph}
\end{figure}

\begin{itemize}
    \item \textbf{End-to-end schema-constrained EAE pipeline}. We formulate document level EAE as role-to-span generation per trigger under the official \texttt{MAVEN-ARG} JSON output format and evaluator requirements.
    \item \textbf{Controlled comparison of open LLMs.} We benchmark several mid-sized open LLM families under a consistent JSON-only prompting interface, deterministic decoding, and identical evaluation and normalization.
    \item \textbf{Practical schema compliance and output validity}. We use role-set injection and deterministic output validation (JSON extraction, schema filtering, de-duplication, and span alignment) to ensure predictions satisfy strict submission constraints.
    \item \textbf{Empirical gains under official scoring}. Under the official \texttt{MAVEN-ARG} evaluator, fine-tuned open models outperform previously reported few-shot closed-model baselines across mention, entity coreference, and event coreference level evaluations. All code, prompts, and processed data needed to reproduce our experiments are publicly available at \url{https://github.com/dessertlab/EAE/}.
\end{itemize}

\section{Related Work}
\subsection{LLMs and Event Extraction}
\subsubsection{Recent advancements using LLMs.}
Recent work highlights several challenges in event extraction, including the high cost of labeling data and the limited portability of models designed for narrow task formulations. Event mining aims to understand what happened in text and categorize event-related information \cite{Meng2024CEANCE}, but it has historically required substantial labeled supervision to reach strong performance. More recently, large language models (LLMs) such as GPT-style models, LLaMA, and Phi have been applied to event extraction with prompting and instruction-following, enabling more flexible extraction and, in some settings, improved transfer across event types and domains \cite{Yang2024PromptbasedCE}. Overall, this trend reflects a shift from highly task-specific architectures (e.g., BERT-based classifiers or BiLSTM pipelines) toward more general-purpose LLMs that can be adapted to different extraction settings with less bespoke modeling \cite{Meng2024CEANCE}.

A key difficulty, however, is that good performance still depends on carefully controlling model behavior and bridging the gap between open-ended generation and the precise requirements of event extraction (e.g., strict schemas and span-faithful outputs). Moreover, results can remain sensitive to task design choices such as prompt format, context selection, and output constraints \cite{sivarajkumar2023empiricalevaluationpromptingstrategies}.

\subsubsection{Event Extraction methods.}
LLM-based event extraction has been studied through several increasingly common approaches, including prompt engineering, few-shot learning, zero-shot learning, and retrieval-augmented generation. Prompt engineering focuses on designing instructions and templates that guide the model toward correct structured predictions in zero-shot and few-shot settings, often with minimal additional training \cite{sivarajkumar2023empiricalevaluationpromptingstrategies}. Few-shot learning uses a small number of labeled examples and can be useful when data is sparse, where demonstration selection and template design play an important role. Zero-shot learning goes a step further by aiming to identify event types and roles without labeled examples, often by aligning contexts with label descriptions or external knowledge; it is also useful for cross-lingual settings, where language-agnostic prompts can support role and argument identification \cite{yue-etal-2023-zero,zhang2022efficientzeroshoteventextraction}. Retrieval-augmented generation can further improve robustness by retrieving relevant examples or auxiliary context at inference time, which can increase accuracy when the retrieved information is well-aligned with the target schema and format requirements.

Across these methods, a persistent challenge is that improvements often hinge on prompt design, the context provided to the model, and whether the model’s generation behavior aligns with the extraction template and schema constraints \cite{sivarajkumar2023empiricalevaluationpromptingstrategies}. 
In addition, comparisons across LLM families (e.g., GPT-style models, Gemini, and LLaMA-2) can be confounded by prompt, context-length limits, and decoding settings, so cross-model results should be interpreted cautiously without a controlled protocol \cite{geminiteam2025geminifamilyhighlycapable,touvron2023llama2openfoundation}.

\paragraph{Our approach.}
While prior work explores prompting, few/zero-shot learning, and augmentation for event extraction, it remains unclear how well open LLMs perform on \emph{document-level, schema-constrained} EAE when outputs must satisfy a strict JSON interface and are scored by the official MAVEN-ARG evaluator. Existing results are often reported under different prompts, context windows, and decoding settings, making comparisons difficult. In contrast, we provide a unified pipeline -- role-set conditioned prompting, parameter-efficient supervised fine-tuning, deterministic decoding, and schema-aware output validation -- and benchmark multiple \textit{mid-sized open} LLMs under a consistent protocol. Our best fine-tuned open model ({Phi-4, 14B}) 
achieves {42.39} event-coreference F1, significantly outperforms the strongest reported GPT baseline in event-coreference F1 (GPT-4) by +12.89 absolute points under the official MAVEN-ARG evaluator. This shows that that lightweight open models, could outperform few-shot GPT prompting while enabling significantly low cost local inference on a single-GPU with 4-bit base weights.

\subsection{Datasets and Benchmarks}
\subsubsection{Key datasets for event extraction.}
ACE 2005 \cite{walker2006ace2005multilingual} is one of the most widely used datasets for event extraction and includes documents in English, Arabic, and Chinese, with annotated event types, subtypes, and argument roles. It has been widely used for trigger identification and argument classification, and it is frequently used as a reference point in event extraction studies \cite{liu2021overvieweventextractionapplications,deng-etal-2022-title2event}. DEIE is a large document-level dataset that includes triggers, arguments, summaries, and relationships, and supports broader document-level event information extraction \cite{ren-etal-2024-deie}. Title2Event focuses on event extraction from news titles at scale \cite{deng-etal-2022-title2event}.

In this work, MAVEN-ARG is our main benchmark. It provides a large schema with many event types and argument roles and includes expert-written definitions as well as annotations that support argument and coreference-aware evaluation \cite{wang-etal-2024-maven}. The dataset and the official evaluation toolkit are publicly released by the MAVEN-ARG benchmark authors and available through the official repository.\footnote{\url{https://github.com/THU-KEG/MAVEN-Argument}} Because MAVEN-ARG evaluation requires strict structured outputs and coreference-aware aggregation, it provides a stringent testbed for studying schema compliance and output validity in LLM-based EAE. Prior results on MAVEN-ARG indicate that stronger closed LLMs (e.g., GPT-4) can outperform weaker closed baselines (e.g., GPT-3.5) across mention-, entity-coreference-, and event-coreference-based evaluation levels under the official scorer \cite{wang-etal-2024-maven}. These characteristics make MAVEN-ARG a challenging benchmark for document-level argument extraction under strict schema and evaluation constraints.

Benchmark results on ACE 2005 are also commonly used for comparison. For example, ChatGPT has been evaluated for event extraction in dialogue-oriented settings and for data generation/augmentation, with reported improvements including better coverage of long-tail event types \cite{kan-etal-2024-emancipating}. However, ACE 2005 contains more entity-centered arguments and a more limited ontology than MAVEN-ARG; as a result, it is less comprehensive for studying document-level argument extraction with coreference-aware evaluation \cite{kan-etal-2024-emancipating}.

\section{Method}

This paper evaluates LLMs for \emph{document-level} event argument extraction under \emph{strict schema} and \emph{structured-output} constraints. We follow the \texttt{MAVEN-ARG} setting, where each instance provides a document, a target event type, and a specific trigger mention. Given these inputs, the system must extract argument spans from the document and assign each span to a role that is valid for the target event type.
Our method is motivated by two recurring failure modes observed in LLM-based EAE. The first is \textbf{schema inconsistency}: models may hallucinate roles that are invalid for the given event type or introduce extra keys that are not part of the allowed role set. This behavior reduces accuracy and can also break the strict interface expected by the official evaluator. To address it, we explicitly inject the event-specific role set into every prompt and apply schema filtering during post-processing to remove any out-of-schema keys.

\begin{figure}[t]
\centering
\scriptsize
\resizebox{\linewidth}{!}{%
\begin{tikzpicture}[
  >=Latex,
  line/.style={-Latex, thick},
  box/.style={draw, rounded corners, align=center, inner sep=3pt, minimum height=8mm},
  group/.style={draw, rounded corners, dashed, inner sep=2pt}
]

\node[box, text width=2.4cm] (train) {Train/Valid\\MAVEN-ARG};
\node[box, text width=3.2cm, right=6mm of train] (prompt_tr) {Task formulation\\+ roleset conditioning };
\node[box, text width=2.6cm, right=6mm of prompt_tr] (sft) {LoRA SFT \\+Save adapters (\texttt{unsloth})};

\draw[line] (train) -- (prompt_tr);
\draw[line] (prompt_tr) -- (sft);

\node[group, fit=(train) (sft), label=above:{\textbf{Training}}] {};

\node[box, text width=3.6cm, below=10mm of train] (test)
{Test instance\\(doc, event type, trigger)\\+ roleset-injected prompt};
\node[box, text width=2.6cm, right=6mm of test] (infer)
{Deterministic\\inference\\(greedy, EOS stop)};
\node[box, text width=2.9cm, right=6mm of infer] (post)
{Schema-aware\\post-processing\\(parse/filter/dedupe)};
\node[box, text width=2.6cm, right=6mm of post] (out)
{Submission-ready\\JSON predictions};

\draw[line] (test) -- (infer);
\draw[line] (infer) -- (post);
\draw[line] (post) -- (out);

\draw[line] (sft.south) to[out=-90,in=90,looseness=1.2] (infer.north);

\node[group, fit=(test) (out), label=above:{\textbf{Inference}}] {};

\end{tikzpicture}%
}
\caption{Overview of our training and inference pipeline for event argument extraction.}
\label{fig:pipeline-overview}
\end{figure}
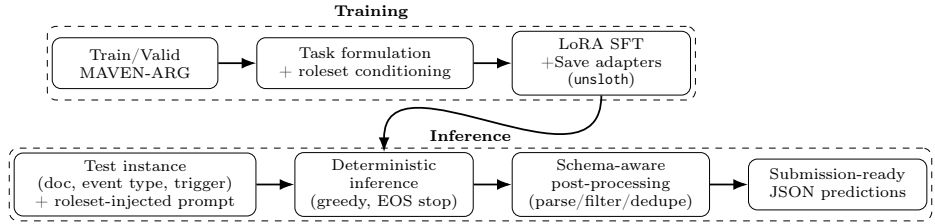

The second failure mode is \textbf{unreliable structured outputs and span faithfulness}. Model generations may contain invalid JSON, inconsistent structures, duplicated values, or argument strings that do not exactly match the document text. This is particularly damaging in document-level EAE, where a system must produce consistent structured predictions for many triggers in long documents. To mitigate these issues, we use a strict JSON-only instruction format, deterministic decoding, and a post-processing pipeline that validates JSON, normalizes special cases (e.g., \texttt{"NA"}), removes duplicates, and enforces alignment with the schema constraints required by evaluation. Overall, our approach is implemented as a unified pipeline with five stages: task formulation, role-set conditioning, supervised fine-tuning, deterministic inference, and schema-aware post-processing. We describe each stage below. Figure~\ref{fig:pipeline-overview} summarizes our end-to-end pipeline from \texttt{MAVEN-ARG} inputs to submission-ready predictions.

\subsection{Task formulation and output constraints}
We define EAE as a document-level extraction problem constrained by an event schema. The system is given a document $d$, an event type $t \in \mathcal{T}$, and a trigger mention $m$ with character offsets in $d$. The model must produce a single JSON object that maps each valid role $r \in R(t)$ to a list of argument strings extracted from the document text. The role set $R(t)$ is determined by the dataset schema and is treated as a hard constraint at inference time.

The output must contain exactly one JSON object. Each key must be a role name from $R(t)$ and each value must be a list of strings. If no argument is found for a role, the model returns the value \texttt{"NA"}, which is normalized to an empty list during post-processing. 

\subsection{Schema conditioning via roleset induction and prompt design}
\noindent\textbf{Roleset induction for schema control.} To enforce schema compliance, we construct a mapping from event type to its role set. We derive this mapping by scanning gold annotations in the training and development splits and collecting the roles observed for each event type. The resulting mapping is stored in a \texttt{label2role.json} file and is used both to populate the role set in prompts and to filter outputs during post-processing.

\noindent\textbf{Prompt template with roleset injection and trigger marking.} 
All models use a three-part prompt (\texttt{Instruction}/\texttt{Input}/\texttt{Response}), with the role set injected immediately before \texttt{Response}.
To bind extraction to the correct event reference, we explicitly mark the trigger mention in the text window by enclosing the trigger string in \texttt{<< >>}. We use the same prompt format for both supervised fine-tuning and inference; only the \texttt{Response} content differs (gold JSON during training and an empty response during inference).





\begin{figure}[t]
\centering
\begin{mdframed}[innertopmargin=6pt,innerbottommargin=6pt,innerleftmargin=6pt,innerrightmargin=6pt]
\small
\begin{verbatim}
### Instruction:
Extract the list of arguments for the given event type from the
text window.
Use only the roles provided in the Roleset line; do not invent
new roles.
Return ONLY a JSON object where each key is a role and each value
is a list of strings.
If no arguments are found, return NA. Do not include the event
type. Do not add explanations.

### Input:
{"doc_id": "...", "mention_id": "...", "event_type": "...",
 "trigger_char_span": [s,e], "trigger_text": "...",
 "window_offset": [ws,we],
 "text_window": " ... << trigger >> ... "}
Roleset: [Role_1, Role_2, ...]

### Response:
{ ... JSON only ... }
\end{verbatim}
\end{mdframed}
\caption{Three-part prompt template (Instruction / Input / Response) with role-set injection. The trigger is marked with \texttt{<< >>}; at inference the Response section is left empty}
\label{fig:prompt}
\end{figure}

\par\smallskip
This prompt design exposes the model to the allowed role keys for the current event type, reducing role drift and the generation of irrelevant or extra keys. The strict \texttt{Response} header and the JSON-only instruction also reduce free-form continuations and simplify automated parsing.

\subsection{Supervised fine-tuning data construction and LoRA adaptation}

\subsubsection{Building supervised instances per trigger mention.}
For supervised fine-tuning, we create one training instance per trigger mention. For each trigger, we build a text window from the document and highlight the trigger span inside the window. We then collect the gold arguments grouped by role and serialize them into a stable JSON structure that serves as the gold \texttt{Response}. 
Arguments in \texttt{MAVEN-ARG} can refer either to non-entity spans (directly annotated offsets) or to entities with multiple coreferent mentions. For non-entity arguments, we directly extract the text using the annotated offsets. For entity-linked arguments, we select a single surface form from the available entity mentions to include as the argument string. 
To match the local trigger context, we select the entity mention whose midpoint is closest to the trigger midpoint.

\par\medskip
\noindent\textbf{Entity surface selection by trigger proximity.} After collecting arguments for each role, we remove duplicates within each role while preserving order, and serialize the role-to-argument mapping into JSON. This output becomes the gold response for the same training example in the fine-tuning phase.

\par\medskip
\par\medskip

\noindent{\setlength{\fboxsep}{6pt}%
\fbox{%
\begin{minipage}{0.97\linewidth}
\textbf{Input:} entity mentions $\{(s_i,e_i,\textit{text}_i)\}_{i=1}^n$, trigger span $(s_t,e_t)$.\\
Compute $m_t = \frac{s_t + e_t}{2}$ and $m_i = \frac{s_i + e_i}{2}$.\\
Select $j = \arg\min_{i \in \{1,\dots,n\}} |m_i - m_t|$.\\
\textbf{Return:} $\textit{text}_j$.
\end{minipage}%
}}%
\par\medskip
\par\medskip

\noindent\textbf{Parameter-efficient fine-tuning with Unsloth and LoRA.}
{\sloppy
We fine-tune open-source models using the \textsc{Unsloth} framework with parameter-efficient LoRA adapters. We apply LoRA to both attention projection layers and feed-forward submodules in each transformer block, including \texttt{q\_proj}, \texttt{k\_proj}, \texttt{v\_proj}, \texttt{o\_proj}, \texttt{gate\_proj}, \texttt{up\_proj}, and \texttt{down\_proj}. We enable gradient checkpointing for memory efficiency and load base model weights in 4-bit precision to support single-GPU training.
\par}
A key aspect of our setup is \emph{interface alignment}: fine-tuning uses exactly the same strict prompt format used at inference time, including role-set injection and JSON-only outputs. This reduces train--test mismatch and makes the model less likely to deviate from the required structured output format at test time.

\noindent\textbf{Deterministic inference and prompt preservation.}

During inference, we use greedy decoding (temperature $=0$) to obtain deterministic and repeatable outputs, enabling more controlled comparisons across different model backbones. We configure the tokenizer with left truncation and left padding so that, for long text windows, the \texttt{Response} header remains close to the end of the input sequence. Preserving this response anchor is important because the model must reliably start generating immediately after the \texttt{Response} header and produce JSON only.

Inference prompts are constructed identically to training prompts, except that the \texttt{Response} section is left empty. The role set is injected from \texttt{label2role.json} for each event type, ensuring that the model always receives explicit schema constraints.

\noindent\textbf{Post-processing, schema validation, and submission formatting.}
After generation, predictions are post-processed to ensure they are valid, schema-consistent, and compatible with the benchmark submission format. This step is necessary because even strong models may produce formatting errors or out-of-schema keys that would otherwise prevent evaluation.

The post-processing pipeline includes:
\begin{itemize}
  \item \textbf{JSON extraction and validation.} We parse the output as JSON; we do not attempt to repair malformed generations, and any parsing failure is discarded and treated as an empty prediction.
  \item \textbf{Normalization of missing values.} We normalize \texttt{"NA"} to an empty list to keep value types consistent.
  \item \textbf{Event-type attachment.} 
  We attach the corresponding \texttt{event\_type} 
  using an auxiliary mapping to match the benchmark submission format.
  \item \textbf{Schema filtering.} We remove any keys that are not in the injected role set. 
  \item \textbf{De-duplication and merging.} Within each role, we remove duplicate argument strings; if multiple generations produce overlapping outputs for the same reference, we merge them.
  \item \textbf{Window alignment.} We verify that each predicted span occurs within the provided text window; spans not found in the window are discarded.
\end{itemize}



\begin{figure*}[t]
\centering
\resizebox{\textwidth}{!}{%
\begin{tikzpicture}[x=1cm,y=1cm,>=Latex]

\node[font=\itshape\large] at (4.4,9.0) {Source document schema (one dataset record)};
\node[font=\itshape\large] at (13.6,9.0) {Expected inference output schema};
\node[
    fill=panelbg,
    rounded corners=2pt,
    minimum width=8.8cm,
    minimum height=9.0cm,
    anchor=north west
] (Lbox) at (0,8.45) {};

\node[
    fill=panelbg,
    rounded corners=2pt,
    minimum width=8.8cm,
    minimum height=9.0cm,
    anchor=north west
] (Rbox) at (9.2,8.45) {};

\begin{scope}[shift={(0.18,8.24)}, yscale=1.12]
\tikzset{every node/.style={anchor=north west,font=\ttfamily\scriptsize,inner sep=0pt,outer sep=0pt}}

\node (l1)  at (0,0.00)  {\jk{\{ "id": } \js{"364ed14fc610df6e25a2f446e2b2d2ab"}\jk{,}};
\node (l2)  at (0,-0.22) {\jk{  "title": } \js{"Expedition of the Thousand"}\jk{,}};

\node (l3)  at (0,-0.44) {\jk{  "document": } \js{"The Expedition of the Thousand (Italian ``Spedizione dei}};
\node (l4)  at (0,-0.66) {\js{ Mille'') was an event of the Italian Risorgimento that took place in}};
\node (l5)  at (0,-0.88) {\js{ 1860. A corps of volunteers led by Giuseppe Garibaldi sailed from}};
\node (l6)  at (0,-1.10) {\js{ Quarto, near Genoa (now Quarto dei Mille), ...."}\jk{,}};

\node (l7)  at (0,-1.34) {\jk{  "events": [}};
\node (l8)  at (0.35,-1.56) {\jk{\{ "id": } \js{"EVENT\_451b7cde13d2b8c21426db027c51096f"}\jk{,}};
\node (l9)  at (0.70,-1.78) {\jk{  "type": } \js{"Motion"}\jk{,}};
\node (l10) at (0.70,-2.00) {\jk{  "type\_id": } \jn{46}\jk{,}};
\node (l11) at (0.70,-2.22) {\jk{  "mention": [}};
\node (l12) at (1.05,-2.44) {\jk{\{ "id": } \js{"1c6807377cd111b86d2d3afc85bf2ca2"}\jk{,}};

\node (l13) at (1.40,-2.72) {\jk{  "trigger\_word": } \js{"sailed"}\jk{,}};
\node (l14) at (1.40,-3.00) {\jk{  "offset": } \jn{[185, 191]}};
\node (l15) at (1.05,-3.28) {\jk{\}}};
\node (l16) at (0.70,-3.50) {\jk{  ],}};
\node (l17) at (0.70,-3.72) {\jk{  "argument": \{}};

\node (l18) at (1.05,-3.94) {\jk{    "Agent": [}};
\node (l19) at (1.40,-4.16) {\jk{\{ "content": } \js{"a corps of volunteers led by Giuseppe}};
\node (l20) at (1.40,-4.38) {\js{ Garibaldi"}\jk{,}};
\node (l21) at (1.75,-4.60) {\jk{        "offset": } \jn{[137, 184]}};
\node (l22) at (1.40,-4.82) {\jk{      \}}};
\node (l23) at (1.05,-5.04) {\jk{    ],}};

\node (l24) at (1.05,-5.26) {\jk{    "Location\_original": [}};
\node (l25) at (1.40,-5.48) {\jk{\{ "entity\_id": } \js{"ENTITY\_052c8614c4bb87b5321807d1980332f6"}};
\node (l26) at (1.40,-5.70) {\jk{      \}}};
\node (l27) at (1.05,-5.92) {\jk{    ],}};

\node (l28) at (1.05,-6.14) {\jk{    "Location\_final": [}};
\node (l29) at (1.40,-6.36) {\jk{\{ "entity\_id": } \js{"ENTITY\_185734cbbe4da16fb0782dfaa326529c"}};
\node (l30) at (1.40,-6.58) {\jk{      \}}};
\node (l31) at (1.05,-6.80) {\jk{    ]}};
\node (l32) at (0.70,-7.02) {\jk{  \}}};
\node (l33) at (0.35,-7.24) {\jk{]}};
\node (l34) at (0,-7.46)    {\jk{\}}};

\node[draw=redann, rounded corners=7pt, line width=0.8pt, fit=(l1),  inner xsep=2.4pt, inner ysep=1.6pt] {};
\node[draw=redann, rounded corners=7pt, line width=0.8pt, fit=(l8),  inner xsep=2.4pt, inner ysep=1.6pt] {};
\node[draw=redann, rounded corners=7pt, line width=0.8pt, fit=(l12), inner xsep=2.4pt, inner ysep=1.6pt] {};
\node[draw=redann, rounded corners=7pt, line width=0.8pt, fit=(l13), inner xsep=2.4pt, inner ysep=1.6pt] {};
\node[draw=redann, rounded corners=7pt, line width=0.8pt, fit=(l14), inner xsep=2.4pt, inner ysep=1.6pt] {};
\node[draw=redann, rounded corners=7pt, line width=0.8pt, fit=(l21), inner xsep=2.4pt, inner ysep=1.6pt] {};

\node[draw=blueann, rounded corners=7pt, line width=0.8pt, fit=(l18), inner xsep=3.2pt, inner ysep=2.0pt] {};
\node[draw=blueann, rounded corners=7pt, line width=0.8pt, fit=(l24), inner xsep=3.2pt, inner ysep=2.0pt] {};
\node[draw=blueann, rounded corners=7pt, line width=0.8pt, fit=(l28), inner xsep=3.2pt, inner ysep=2.0pt] {};
\end{scope}

\begin{scope}[shift={(9.38,8.24)}, yscale=1.12]
\tikzset{every node/.style={anchor=north west,font=\ttfamily\scriptsize,inner sep=0pt,outer sep=0pt}}

\node (r1)  at (0,0.00)  {\jk{\{}};
\node (r2)  at (0,-0.22) {\jk{  "id": } \js{"364ed14fc610df6e25a2f446e2b2d2ab"}\jk{,}};
\node (r3)  at (0,-0.44) {\jk{  "preds": \{}};

\node (r4)  at (0.35,-0.70) {\jk{    "1c6807377cd111b86d2d3afc85bf2ca2": \{}};
\node (r5)  at (0.70,-0.96) {\jk{      "event\_type": } \js{"Motion"}\jk{,}};
\node (r6)  at (0.70,-1.22) {\jk{      "Agent": [}};
\node (r7)  at (1.05,-1.48) {\js{"a corps of volunteers led by}};
\node (r8)  at (1.05,-1.74) {\js{Giuseppe Garibaldi"}};
\node (r9)  at (0.70,-2.00) {\jk{      ],}};
\node (r10) at (0.70,-2.26) {\jk{      "Location\_original": [}};
\node (r11) at (1.05,-2.52) {\js{"Quarto dei Mille"}};
\node (r12) at (0.70,-2.78) {\jk{      ],}};
\node (r13) at (0.70,-3.04) {\jk{      "Location\_final": [}};
\node (r14) at (1.05,-3.30) {\js{"Marsala, Sicily"}};
\node (r15) at (0.70,-3.56) {\jk{      ]}};
\node (r16) at (0.35,-3.82) {\jk{    \},}};

\node (r17) at (0.35,-4.08) {\jk{    "187955a9227516bae5cf7198e30a3cd4": \{}};
\node (r18) at (0.70,-4.30) {\jk{      "event\_type": } \js{"Motion"}\jk{,}};
\node (r19) at (0.70,-4.52) {\jk{      "Agent": [}};
\node (r20) at (1.05,-4.74) {\js{"a corps of volunteers led by}};
\node (r21) at (1.05,-4.96) {\js{Giuseppe Garibaldi"}};
\node (r22) at (0.70,-5.18) {\jk{      ],}};
\node (r23) at (0.70,-5.40) {\jk{      "Location\_original": [}};
\node (r24) at (1.05,-5.62) {\js{"Quarto, near Genoa"}};
\node (r25) at (0.70,-5.84) {\jk{      ],}};
\node (r26) at (0.70,-6.06) {\jk{      "Location\_final": [}};
\node (r27) at (1.05,-6.28) {\js{"Marsala, Sicily"}};
\node (r28) at (0.70,-6.50) {\jk{      ]}};
\node (r29) at (0.35,-6.72) {\jk{    \}}};
\node (r30) at (0,-6.94) {\jk{  \}}};
\node (r31) at (0,-7.16) {\jk{\}}};

\node[draw=redann, rounded corners=7pt, line width=0.8pt, fit=(r2), inner xsep=4pt, inner ysep=2.4pt] {};
\node[draw=redann, rounded corners=7pt, line width=0.8pt, fit=(r4), inner xsep=4pt, inner ysep=2.4pt] {};

\node[draw=blueann, rounded corners=7pt, line width=0.8pt, fit=(r6),  inner xsep=4pt, inner ysep=2.4pt] {};
\node[draw=blueann, rounded corners=7pt, line width=0.8pt, fit=(r10), inner xsep=4pt, inner ysep=2.4pt] {};
\node[draw=blueann, rounded corners=7pt, line width=0.8pt, fit=(r13), inner xsep=4pt, inner ysep=2.4pt] {};
\end{scope}

\node[draw, minimum width=2.0cm, minimum height=0.8cm, font=\normalsize] (llm) at (9.0,-1.10) {Open LLMs};
\draw[line width=0.8pt] (2.9,-0.55) -- (2.9,-0.85);
\draw[line width=0.8pt] (16.1,-0.55) -- (16.1,-0.85);
\draw[-{Latex[length=2.5mm]}, line width=0.8pt] (2.9,-0.85) -- (llm.west);
\draw[-{Latex[length=2.5mm]}, line width=0.8pt] (llm.east) -- (16.1,-0.85);
\node[font=\normalsize] at (5.45,-1.25) {Input};
\node[font=\normalsize] at (12.85,-1.25) {Expected Output};

\end{tikzpicture}%
}
\caption{Source document schema and expected inference output schema.
Red boxes highlight metadata and trigger fields (e.g.\ \texttt{id},
\texttt{trigger\_word}, \texttt{offset}); blue boxes highlight the
schema argument roles (e.g.\ \texttt{Agent}, \texttt{Location\_original},
\texttt{Location\_final})}
\label{fig:database-schema-internal-mapping}
\end{figure*}


As shown in Figure~\ref{fig:database-schema-internal-mapping}, post-processing enforces schema consistency and structured-output validity, reducing preventable errors such as malformed JSON, redundant or irrelevant keys, and duplicated values. As a result, the system produces structured JSON outputs in the exact format required by the official evaluator.

\section{Evaluation}

This section describes the experimental setup used to evaluate LLMs for document level event argument extraction (EAE) on MAVEN-ARG. We summarize the dataset and its schema, explain how entity and event coreference affect evaluation, detail the official scoring and normalization rules, and present the compared models and baselines. Our main research question is whether mid-sized open LLMs, when adapted via supervised fine-tuning and constrained by schema-conditioned prompts, can match or surpass previously reported few-shot GPT baselines on MAVEN-ARG under the official evaluation protocol, and how performance changes across mention-, entity-coreference-, and event-coreference-based evaluation levels.

\subsection{Dataset and dataset schema}

In all experiments, we use MAVEN-ARG \cite{wang-etal-2024-maven}, a document-level EAE dataset derived from MAVEN and built from English Wikipedia articles. MAVEN-ARG provides full argument annotations for both entity-linked and non-entity spans and is designed to support cross-sentence extraction with full-document context. The dataset defines an extensive nested JSON schema with 162 event types and 612 argument roles and includes expert-written definitions and human annotations aligned to character offsets.

The training and development splits contain full document annotations. Each document provides raw text, a list of events, and a list of entities. Each event specifies an event type and contains one or more trigger mentions with offsets, along with an argument dictionary whose keys are roles. Arguments are represented in two forms: (i) entity-linked arguments that refer to an entity identifier associated with a within-document coreference cluster, and (ii) non-entity arguments recorded as free spans with surface text and character offsets. The test split preserves blind evaluation: it provides event and entity references, but gold arguments and coreference links required for scoring are kept private, and evaluation is performed by submitting predictions in the required JSON format to the official online platform. 
We follow the official split with 2{,}913 training documents, 710 development documents, and 857 test documents, containing 70{,}775, 16{,}996, and 19{,}736 events, respectively \cite{wang-etal-2024-maven}.

\subsection{Coreference annotations and evaluation metrics and levels}

MAVEN-ARG includes entity and event coreference: entities form mention clusters and events form trigger clusters referring to the same underlying event \cite{wang-etal-2024-maven}, enabling evaluation beyond strict mention matching.

We report results at the three levels supported by the official scorer. \textbf{Mention-level} evaluation scores each trigger mention independently by comparing its predicted role-to-argument mapping to the gold mapping. \textbf{Entity-coreference} evaluation relaxes surface-form matching: a prediction can be counted as correct if it matches any coreferent mention of the gold entity, reducing sensitivity to which mention string is produced. \textbf{Event-coreference} evaluation aggregates trigger mentions that belong to the same event cluster and evaluates arguments at the cluster level, rewarding systems that retrieve correct arguments consistently across multiple coreferent triggers. Together, these levels reflect the document-level nature of MAVEN-ARG and measure robustness to coreference and cross-sentence dependencies.

\subsection{Official metrics and normalization}

All reported scores follow the official MAVEN-ARG evaluation procedure \cite{wang-etal-2024-maven}. We report Precision, Recall, F1, and Exact Match at mention, entity-coreference, and event-coreference levels. Prior to scoring, both gold and predicted argument strings are normalized by lowercasing, removing punctuation, and collapsing extra whitespace. Exact Match requires the normalized predicted string to exactly equal the normalized gold string. In addition to Exact Match, the scorer computes bag-of-words similarity based on word overlap.

At the mention level, scoring uses role-wise bipartite matching between the predicted and gold argument lists. For each role, the scorer constructs a similarity matrix between items in the gold and predicted lists and selects the best one-to-one matching to maximize the total similarity. The scorer also applies a cardinality penalty when the numbers of predicted and gold arguments differ, penalizing over- and under-generation. Scoring is schema-aware: only roles valid for the gold event type can contribute positively, and out-of-schema role keys do not yield credit. Entity-coreference and event-coreference scoring extend this procedure by allowing matches against coreferent entity mentions and by aggregating across coreferent trigger mentions, respectively.

\subsection{Models and baselines}

We evaluate several mid-sized open LLMs using supervised fine-tuning with the schema-conditioned prompting interface described in Section~3. The open models include Llama-3.1 (8B), Mistral-Nemo (12B), Qwen3 (14B), and Phi-4 (14B). For all open models, we inject the event-specific role set into the prompt, require JSON-only outputs, decode deterministically, and apply the same validation and schema-filtering post-processing prior to evaluation. At inference the \texttt{Response} section is left empty, the role set appears only in the \texttt{Roleset} line of the prompt, and we do not prefill the answer keys.

As baselines, we report GPT-3.5 and GPT-4 results previously reported on MAVEN-ARG using two-shot in-context learning, both with and without adding event/role definitions to the prompt \cite{wang-etal-2024-maven}. These baselines serve as reference points for closed-model few-shot prompting without supervised fine-tuning on MAVEN-ARG. We do not additionally evaluate the open models under few-shot prompting, since our scope is the fine-tuned schema-conditioned setting, and few-shot evaluation of open models is left to future work. Since we do not re-run these systems, differences in prompt templates, context-length limits, and decoding settings across studies may affect strict comparability, so we interpret comparisons to published closed-model results as indicative rather than strictly controlled, and we do not claim a fully controlled head-to-head comparison against closed models.

\subsection{Compute environment and reproducibility protocol}

Training and inference are performed on an NVIDIA Tesla T4 GPU (16 GB VRAM) with CUDA support. To fit models within hardware constraints, base weights are loaded with 4-bit quantization during fine-tuning and inference. Our implementation uses Python 3.12 and PyTorch 2.8.0 (CUDA), along with standard libraries for training and inference, including \texttt{transformers}, \texttt{trl}, \texttt{peft}, and \texttt{unsloth}.

We keep the testing protocol constant across all open models. We construct the event-type-to-role-set mapping from the training and development splits, build supervised instances at the trigger-mention level using the same strict Instruction/Input/Response format with role-set injection, and decode deterministically. Generated outputs are post-processed via a JSON extraction pipeline, normalization, and schema filtering to produce predictions in the required format by the official evaluator. Final test scores are obtained from the official MAVEN-ARG online evaluation platform using the required submission format.

\subsection{Implementation details}
We perform supervised fine-tuning with LoRA using the \texttt{unsloth} training stack. Unless otherwise specified, we use a maximum input length of \texttt{2048} tokens and a maximum generation length of  \texttt{512} new tokens at inference time. For sequences longer than \texttt{2048} tokens, we rely on standard tokenizer truncation to the maximum sequence length. We decode deterministically using greedy decoding with \texttt{temperature $=0$}, \texttt{top-$p$=1.0}, a repetition penalty of \texttt{1.05} and stop at the end-of-turn token $\mathtt{<|im\_end|>}$ (EOS).

LoRA adapters use rank $r=128$, $\alpha=32$, and dropout $=0$, applied to \texttt{q\_proj,k\_proj,v\_proj,o\_proj,gate\_proj,up\_proj,down\_proj}. We train with AdamW \texttt{adamw\_8bit} at a learning rate \texttt{2e-4}, batch size \texttt{8} with gradient accumulation \texttt{4}, for \texttt{60} optimization steps with linear scheduling, with warmup ratio \texttt{5} warmup steps and weight decay \texttt{0.01}. We use BF16 when supported (otherwise FP16), gradient checkpointing, a max grad norm of \texttt{1.0} and load base weights in 4-bit quantization on a single NVIDIA Tesla T4 GPU (16 GB VRAM). We note that this corresponds to a low-data regime, since at batch size \texttt{8} with gradient accumulation \texttt{4} over \texttt{60} steps, fine-tuning uses roughly \texttt{1{,}920} trigger-level instances, on the order of \texttt{3\%} of the \texttt{70{,}775} training events. Further details regarding the implementation and experimental setup can be found in the accompanying GitHub repository.


\section{Results and Discussion}
In this section, we report results on the MAVEN-ARG test set using the official evaluator and discuss the main performance patterns observed across models. We compare mid-sized open LLMs fine-tuned with LoRA under the schema-conditioned prompting interface (Section~3) against previously reported GPT-3.5 and GPT-4 baselines on MAVEN-ARG. We evaluate at three official levels: \textbf{mention}, \textbf{entity coreference}, and \textbf{event coreference}. All metrics are micro-averaged \textbf{Precision}, \textbf{Recall}, \textbf{F1}, and \textbf{Exact Match (EM)}.

\subsection{Overall test results}
Table~\ref{tab:test-set-results-maven-arg-results-split} summarizes official test-set results. Open models are fine-tuned with LoRA using schema-conditioned prompts and JSON-only outputs, while GPT results are reported from prior work as two-shot prompting baselines, both with and without adding role definitions to the prompt \cite{wang-etal-2024-maven}.

\begin{table}[t]
\caption{Test-set results on MAVEN-ARG. All numbers are percentages.}
\label{tab:test-set-results-maven-arg-results-split}
\centering

\renewcommand{\arraystretch}{1.05}

\textbf{Mention Level}\par
\vspace{2pt}
\setlength{\tabcolsep}{4pt}
\footnotesize{\begin{tabular}{lcccc}
\toprule
Model & P & R & F1 & EM \\
\midrule
GPT-3.5               & 21.30 & 20.90 & 19.90 & 14.30 \\
GPT-3.5 + definition         & 21.80 & 21.70 & 20.60 & 15.20 \\
GPT-4                 & 25.60 & 27.20 & 25.10 & 17.90 \\
GPT-4 + definition           & 27.20 & 28.70 & 26.60 & 19.10 \\
\midrule
Llama-3.1 (8B)        & 29.62 & 29.73 & 32.27 & 22.79 \\
Mistral-Nemo (12B)    & 34.37 & 31.58 & 31.56 & 24.71 \\
Qwen3 (14B)           & 33.71 & 31.16 & 31.07 & 24.39 \\
\textbf{Phi-4 (14B)}  & \textbf{38.61} & \textbf{36.22} & \textbf{35.99} & \textbf{29.28} \\
\bottomrule
\end{tabular}}

\vspace{16pt}

\textbf{Entity Coref Level and Event Coref Level}\par
\vspace{2pt}
\setlength{\tabcolsep}{3pt}
\footnotesize{\begin{tabular}{lcccccccc}
\toprule
\multirow{2}{*}{Model} &
\multicolumn{4}{c}{Entity Coref Level} &
\multicolumn{4}{c}{Event Coref Level} \\
\cmidrule(lr){2-5}\cmidrule(lr){6-9}
 & P & R & F1 & EM & P & R & F1 & EM \\
\midrule
GPT-3.5               & 24.50 & 25.10 & 23.40 & 16.80 & 24.40 & 24.80 & 23.20 & 16.90 \\
GPT-3.5 + def         & 25.00 & 25.80 & 24.10 & 17.80 & 24.90 & 25.40 & 23.90 & 17.90 \\
GPT-4                 & 28.90 & 31.70 & 28.70 & 20.20 & 27.90 & 30.50 & 27.60 & 19.50 \\
GPT-4 + def           & 30.50 & 33.30 & 30.30 & 21.30 & 29.80 & 32.30 & 29.50 & 21.10 \\
\midrule
Llama-3.1 (8B)        & 35.47 & 35.80 & 38.09 & 28.14 & 37.73 & 35.48 & 35.14 & 27.82 \\
Mistral-Nemo (12B)    & 40.70 & 38.10 & 37.90 & 30.58 & 40.21 & 37.71 & 37.48 & 30.18 \\
Qwen3 (14B)           & 39.58 & 37.30 & 36.97 & 29.80 & 29.77 & 39.48 & 37.26 & \textbf{36.92} \\
\textbf{Phi-4 (14B)}  & \textbf{46.34} & \textbf{44.08} & \textbf{43.64} & \textbf{36.26} & \textbf{45.07} & \textbf{42.75} & \textbf{42.39} & 35.18 \\
\bottomrule
\end{tabular}}
\end{table}

Across all three evaluation levels, fine-tuned open LLMs substantially outperform the reported few-shot GPT baselines. Among open models, \textbf{Phi-4 (14B)} is the most robust overall, achieving the best event-coreference F1 (42.39\%). Mistral-Nemo and Qwen3 also perform strongly at the event-coreference level, followed by Llama-3.1.  
Notably, a mid-sized open model (Phi-4, 14B) achieves the best event-coreference F1, outperforming the published GPT baselines reported for MAVEN-ARG under the official evaluator. 

We observe that F1 scores are consistently higher than Exact Match, indicating that many predictions are \emph{near matches} to the reference strings (high token overlap) but do not reproduce the gold span exactly. Finally, note that the official scorer uses Hungarian (bipartite) matching with a cardinality penalty, so the reported micro-averaged F1 is not guaranteed to equal the harmonic mean of the displayed global precision and recall and can exceed both.

\subsection{Comparison against GPT baselines}

\begin{figure}[t]
\includegraphics[width=\textwidth]{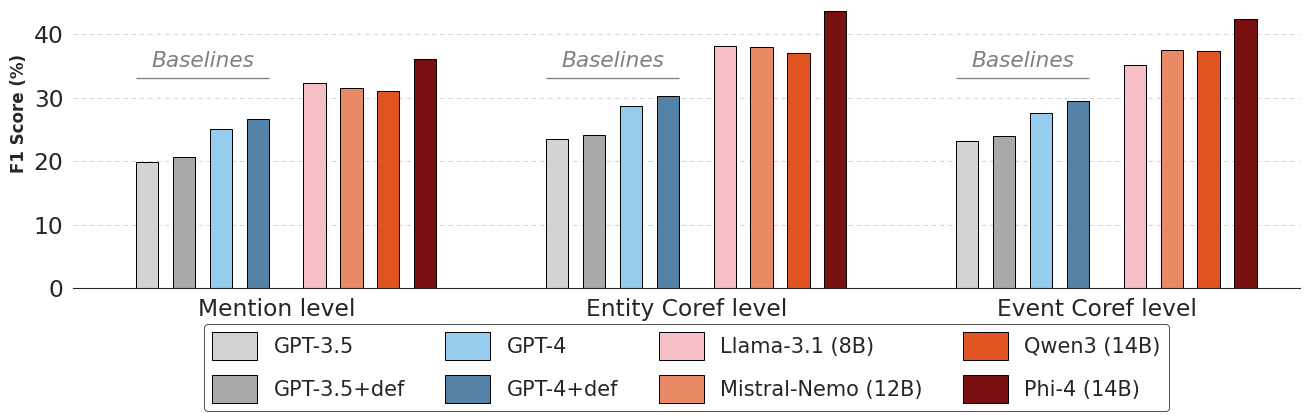}
\caption{F1 performance on MAVEN-ARG: GPT baselines vs. Open LLMs} \label{fig:F1-score-performance}
\end{figure}

Figure~\ref{fig:F1-score-performance} highlights event-coreference F1 gains over the strongest reported GPT baseline (GPT-4 + definition). All fine-tuned open LLMs show large improvements under the official evaluator, suggesting that supervised adaptation with a strict schema-conditioned interface is more effective than few-shot prompting alone for this benchmark.

\subsection{Precision--recall profiles at the event-coreference level}
Models show different precision--recall trade-offs at the event-coreference level. Phi-4, Mistral-Nemo, and Llama-3.1 are more precision-oriented, indicating more conservative extraction (fewer spurious arguments). 
In contrast, Qwen3 is recall-biased, extracting more candidate arguments and thus more false positives, consistent with its precision--recall gap in Table~\ref{tab:test-set-results-maven-arg-results-split}.

\subsection{Exact Match versus near matches}
The gap between bag-of-words F1 and Exact Match indicates that span boundaries and surface-form fidelity remain important sources of error. At entity- and event-coreference levels, Llama-3.1, Mistral-Nemo, and Phi-4 show clear F1--EM gaps: many predictions are close to the reference (earning token-overlap credit) but fail Exact Match due to boundary differences, modifiers, or small surface variations. In contrast, Qwen3 shows a smaller F1--EM gap at the event-coreference level; its main weakness is precision rather than boundary fidelity, consistent with the model producing additional plausible but non-gold arguments.

\subsection{Qualitative case studies}
\begin{figure}[t]
\includegraphics[width=\textwidth]{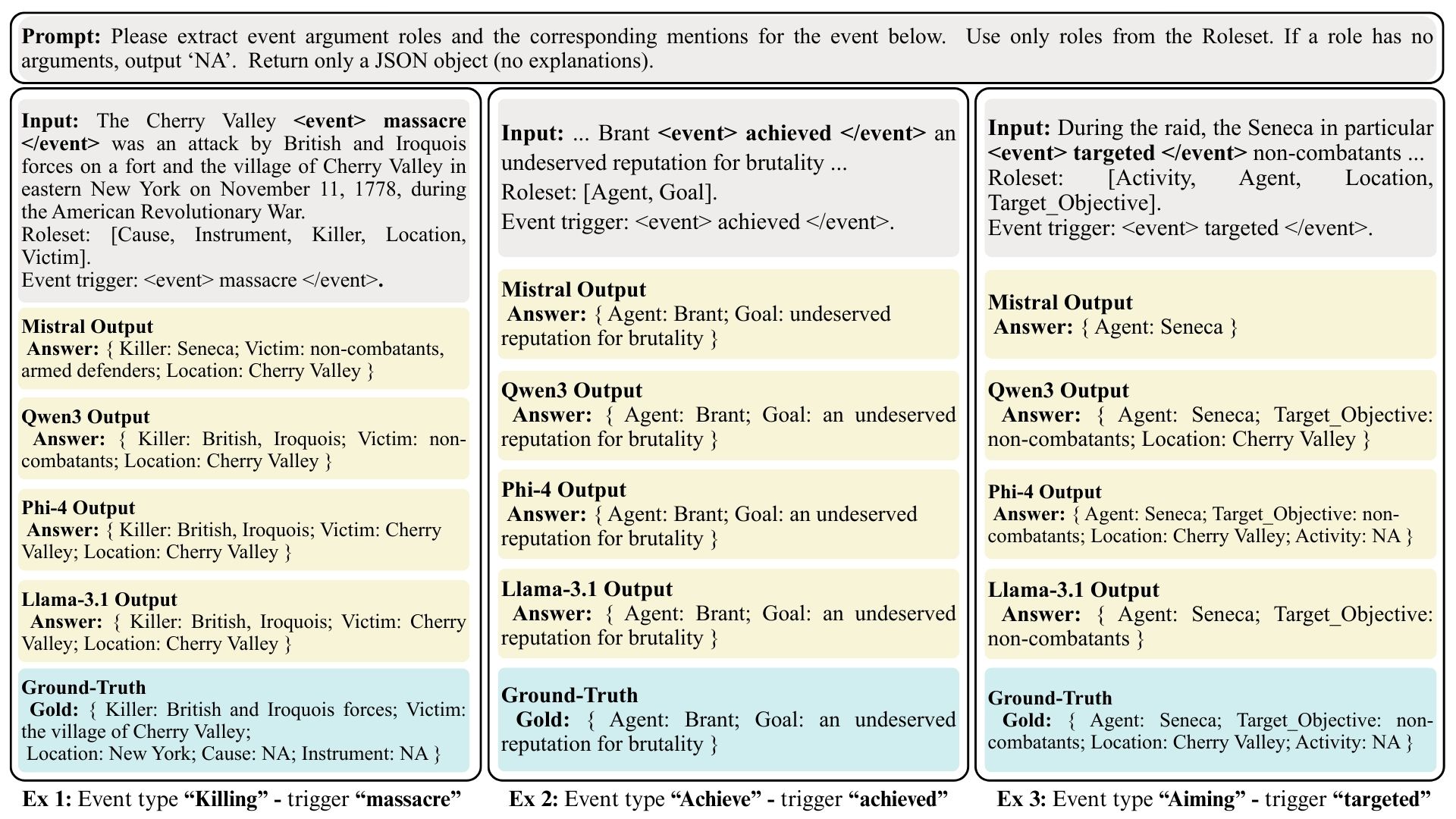}
\caption{Qualitative examples of event argument extraction} \label{fig:qualitative-examples}
\end{figure}

To connect metrics to model behavior, we analyze the qualitative examples in Figure~\ref{fig:qualitative-examples}. In the \textsc{Killing} example (trigger: ``massacre''), systems often identify correct roles but differ in span selection, sometimes extracting shorter spans or near matches that are not faithful to the gold text. In the \textsc{Achieve} example (trigger: ``achieved''), models more consistently identify \textsc{Agent} and \textsc{Goal}, and predictions tend to be closer to reference strings. In the \textsc{Aiming} example (trigger: ``targeted''), systems differ in role coverage: some models output many roles from the injected role set, while others only output roles they are confident about, illustrating that role coverage remains challenging even with the same text window and schema constraints.

\subsection{Qualitative error signatures from metric patterns}

The metric profiles suggest two dominant error signatures. First, Qwen3 combines high recall with strong EM at the event-coreference level but lower precision, consistent with generating additional plausible arguments that are not present in the gold set. Second, Phi-4, Mistral-Nemo, and Llama-3.1 exhibit larger F1--EM gaps, consistent with near-correct predictions that differ from the gold by span boundaries, extra qualifiers, punctuation, or minor surface-form variations. Token-overlap scoring gives partial credit for these near matches, while Exact Match penalizes them.

\subsection{Key takeaways}

Overall, \textbf{Phi-4 (14B)} achieves the best performance at the event-coreference level under the official evaluator. Fine-tuned open LLMs substantially improve over the reported few-shot GPT baselines, benefiting from supervised adaptation and strict schema-conditioned structured outputs. At the same time, models exhibit different precision--recall trade-offs, and the remaining gap between F1 and Exact Match indicates that span boundary fidelity remains a major source of error even when token-overlap scores are high.

\section{Conclusion}
This paper studies document-level event argument extraction with LLMs under strict schema constraints and coreference-aware evaluation. On \texttt{MAVEN-ARG}, we compare previously reported GPT-3.5/GPT-4 two-shot baselines (with/without role definitions) to lightweight open LLMs adapted via supervised fine-tuning. Our pipeline uses role-set conditioned prompting, deterministic decoding, and JSON-validating post-processing. Under the official evaluator, fine-tuned open models consistently outperform the reported GPT baselines at the mention, entity-coreference, and event-coreference levels, showing that mid-sized models can match or exceed published closed-model few-shot results. 
Results also confirm that coreference-aware aggregation improves end-to-end performance, while the gap between bag-of-words F1 and Exact Match indicates that span-boundary fidelity and surface-form variation remain major sources of error.

We envision three directions. First, we will evaluate reasoning-oriented LLMs under the same pipeline and metrics (e.g., stronger Phi variants and DeepSeek-R1~\cite{Guo_2025}) to test whether multi-step reasoning improves span fidelity and role assignment. Second, we will benchmark additional open backbones (for instance, \texttt{gpt-oss-20b}~\cite{openai2025gptoss120bgptoss20bmodel} and newer LLaMA releases) while controlling context length, decoding budget, and instruction-tuning style. Third, we will extend this approach to privacy-sensitive domains such as healthcare, emphasizing local deployment and reproducible workflows aligned with FAIR principles.

\subsubsection{\discintname}
The authors have no competing interests to declare that are
relevant to the content of this article.

\bibliographystyle{splncs04}
\bibliography{references}

\end{document}